\documentclass{article}

\usepackage{microtype}
\usepackage{graphicx}
\usepackage{subfigure}
\usepackage{booktabs} 
\usepackage{mathtools}

\DeclarePairedDelimiter\floor{\lfloor}{\rfloor}

\usepackage{hyperref}

\usepackage[font=small,skip=0pt]{caption}

\usepackage[accepted]{icml2019}

\icmltitlerunning{Towards Interactive Training of Non-Player Characters in Video Games}

\begin{document}

\twocolumn[
\icmltitle{Towards Interactive Training of Non-Player Characters in Video Games}



\icmlsetsymbol{equal}{*}

\begin{icmlauthorlist}
\icmlauthor{Igor Borovikov}{ea}
\icmlauthor{Jesse Harder}{ea}
\icmlauthor{Michael Sadovsky}{sad}
\icmlauthor{Ahmad Beirami}{ea}
\end{icmlauthorlist}

\icmlaffiliation{ea}{Electronic Arts Digital Platform -- Data \& AI, Redwood City, CA, USA}
\icmlaffiliation{sad}{Institute of Computational Modelling, Krasnoyarsk, Russia}

\icmlcorrespondingauthor{Igor Borovikov}{iborovikov@ea.com}
\icmlcorrespondingauthor{Jesse Harder}{jharder@ea.com}
\icmlcorrespondingauthor{Ahmad Beirami}{abeirami@ea.com}

\icmlkeywords{Machine Learning, Human-Computer Interaction, Imitation Learning, Non-Player Characters.}

\vskip 0.3in
]



\printAffiliationsAndNotice{}  

\begin{abstract}
There is a high demand for high-quality Non-Player Characters (NPCs) in video games. Hand-crafting their behavior is a labor intensive and error prone engineering process with limited controls exposed to the game designers. We propose to create such NPC behaviors interactively by training an agent in the target environment using imitation learning with a human in the loop. While traditional behavior cloning may fall short of achieving the desired performance, we show that interactivity can substantially improve it with a modest amount of human efforts. The model we train is a multi-resolution ensemble of Markov models, which can be used as is or can be further ``compressed'' into a more compact model for inference on consumer devices. We illustrate our approach on an example in OpenAI Gym, where a human can help to quickly train an agent with only a handful of interactive demonstrations. We also outline our experiments with NPC training for a first-person shooter game currently in development.
\end{abstract}

\section{Introduction and Motivation}
Autonomous agents in video games are called Non-Player Characters (NPCs). They are an essential element of gameplay in ever increasingly complex game environments. The traditional hand-crafting methods of their creation require a substantial amount of domain knowledge, knowledge engineering, scripting, intuition, and testing. Meanwhile, the requirements for the NPCs are growing in two main dimensions:  scale and depth. The scale calls for various types of characters present in the game to create an illusion of a diversely populated virtual universe. The depth is about making the characters more believable, human-like and engaging. The existing approach to hand-crafting behaviors of NPCs is hard to scale in both dimensions, calling for alternative approaches. The problem of training NPCs in video games may appear highly specific but it shares a lot of challenges with the problem of real-time interactive Machine Learning for fast training and serving custom models in a variety of contexts, e.g., \cite{Crankshaw2015ScalableTA}.

In this paper, we consider an idealized NPC creation workflow, where a game designer  interactively creates NPCs from demonstrations. The designer would produce some examples that leads to training of a model. While the model is running, the designer takes control of the character and produces more demonstrations of the desired behavior until NPCs are trained to the designer's satisfaction. We show that we can make a significant step towards such idealized workflow using simple yet effective techniques based on an ensemble of multi-resolution Markov models.

\section{Related Work}
Using human demonstrations helps training artificial agents in many applications and in particular in video games \cite{Gudmundsson2018HumanLikePW}, \cite{magnus}, \cite{Stanley2006RealTimeIL}. Off-policy human demonstrations are easier to use and are abundant in player telemetry data. Supervised behavior cloning, imitation learning (IL), apprenticeship learning (e.g., \cite{Bogdanovic2015DeepAL}) and generative adversarial imitation learning (GAIL) \cite{Ho2016GenerativeAI} allow for the reproduction of a teacher style and achievement of a reasonable level of performance in the game environment.  Unfortunately, an agent trained using IL is usually unable to effectively generalize to previously underexplored states or to extrapolate stylistic elements of the human player to new states.

Direct inclusion of a human in the control loop can potentially alleviate the problem of limited generalization. Dataset Aggregation, DAGGER \cite{Ross2011ARO}, allows for an effective way of doing that when a human provides consistent optimal input, which may not be realistic in many environments.  Another way of such inclusion of online human input is shared autonomy, which is an active research area with multiple applications, e.g., \cite{Zhou2018SharedAF}, \cite{Rueben2015ASA}, etc. The shared autonomy approach \cite{shared_autonomy_2018} naturally extends to policy blending \cite{Dragan2013APF} and allows to train DQN agents cooperating with a human in complex environments effectively. 
The applications of including a human in the training loop to the fields of robotics and self-driving cars are too numerous to cover here, but they mostly address the optimality aspect of the target policy while here we also aim to preserve stylistic elements of organic human gameplay. 

\section{Markov Ensemble Model}
In this section, we introduce the notation and the necessary background on the building blocks used in the interactive procedure outlined in the next section. 

\textbf{Markov Decision Process:} Following the established treatment of training artificial agents, we place the problem into the framework of Markov Decision Processes (MDPs) \cite{Sutton_Barto}. 
While in many cases the actual observation space may not contain the full state space, the designed algorithms do not distinguish between state $s$ and the corresponding observation available to the agent.
We rely on an extended state space (defined below), which helps to mitigate the partial observability and preserve the stylistic element of human demonstrations. 

The interaction of the agent and the environment takes place at discrete moments $t=1,\dots, T$ with the value of $t$ trivially observable by the agent. The agent, after receiving an observation $s_t$ at time $t$, can take an action $a_t \in A(s, t)$ from the set of allowed actions $A(s, t)$ using policy $\pi: s \to a$. Executing an action in the environment results in a new state $s_{t+1}$ and a reward $r_{t+1}$ also observed by the agent. The primary objective of training a policy in RL is the maximization of cumulative rewards but they are inconsequential for the model we build; hence we drop them from further discussion. In this paper, we consider the episode-based environment, i.e., after reaching a certain condition, the execution of the described state-action loop ends. A complete episode is a sequence $E=\{(s_t, a_t)\}_{t \in {1,\dots,T}}$. The fundamental assumption regarding the described decision process is that it has the Markov property.

\textbf{Extended state:} Besides the most recent action taken before time $t$, i.e., action $a_{t-1}$, we also consider a recent history of the past $n$ actions, where $1 \leq  n < T$, $\alpha_{t, n} := a_{t-n}^{t-1}=\{a_{t-n}, \dots, a_{t-1}\}$, whenever it is defined in an episode $E$. For $n=0$, we define $a_{t,0}$ as the empty sequence. 
We augment the directly observed state $s_t$ with the action history $\alpha_{t,n}$, to obtain an \textit{extended state} $S_{t,n}=(s_t, \alpha_{t,n})$. 

The purpose of including the action history is to better capture additional information (e.g., stylistic features) from human controlling the input during interactive demonstrations. An extended policy $\pi_{n}$, which operates on the extended states $\pi_{n} : S_{t,n} \to a_t$, is useful for modeling human actions in a manner similar to $n$-grams text models in natural language processing (NLP) (e.g., \cite{KaminskiMP}, \cite{davidwrite}, \cite{andresen2017approximating}). Of course, the analogy with $n$-gram models in NLP works only if both state and action spaces are discrete. We will address this restriction in the next subsection using multi-resolution quantization. 

For a discrete state-action space and various $n$, we can compute probabilities $P\{a_t|S_{t,n}\}$ of transitions $S_{t,n} \to a_t$ occurring in (human) demonstrations and use them as a Markov model $M_n$ of order $n$ of (human) actions. We say that the model $M_n$ is defined on an extended state $S_{.,n}$ if the demonstrations contain at least one occurrence of $S_{.,n}$. When a model $M_n$ is defined on $S$, we can use $P\{a_t|S_{t,n}\}$ to sample the next action from all ever observed next actions in state $S_{.,n}$. Hence, $M_n$ defines a partial stochastic mapping $M_n: S_{.,n} \to A$ from extended states to action space $A$.

\textbf{Stacked Markov models:} We call a sequence of Markov models $\mathcal{M}_n = \{M_i\}_{i=0,\dots,n}$ a stack of models. A (partial) policy defined by $\mathcal{M}_n$ computes the next action at a state $s_t$ as described in Algorithm \ref{alg:markov_stack_discrete}. Such policy performs a simple behavior cloning. The policy is partial since it may not be defined on all possible extended states and needs a fallback policy $\pi_*$ to provide a functional agent acting in the environment. 

Note that it is possible to implement sampling from a Markov model using an $\mathcal{O}(1)$ complexity operation with hash tables. Hence, when properly implemented, Algorithm \ref{alg:markov_stack_discrete} is very efficient and suitable for real-time execution in a video game or other interactive application where expected inference time has to be on the scale of 1 ms or less \footnote{A modern video game runs at least at 30 frames per second with lots computations happening during about 33 ms allowed per frame, drastically limiting the ``budget'' allocated for inference.}.

\textbf{Quantization:} Quantization (aka discretization) allows us to work around the limitation of discrete state-action space enabling the application of the Markov Ensemble approach to environments with continuous dimensions. Quantization is commonly used in solving MDPs \cite{RL_state_of_the_art} and has been extensively studied in the signal processing literature \cite{digital_signal_proc}, \cite{vect_quant}. Using quantization schemes that have been optimized for specific objectives can lead to significant gains in model performance, improving various metrics vs. ad-hoc quantization schemes, e.g.,  \cite{RL_state_of_the_art}, \cite{Pages}. 

Instead of trying to pose and solve the problem of optimal quantization, we use a set of quantizers covering a range of schemes from coarse to fine. At the conceptual level, such an approach is similar to multi-resolution methods in image processing, mip-mapping and Level-of-Detail (LoD) representations in computer graphics \cite{comp_graph}. The simplest quantization is a uniform one with step $\sigma$: \[Q_{\sigma}(x) = \sigma \floor*{\frac{x}{\sigma}}\]
For illustration purposes, it is sufficient to consider only the uniform quantization $Q_{\sigma}$. 
In practice, most variables have naturally defined limits which are at least approximately known. Knowing the environment scale gives an estimate of the smallest step size $\sigma_0$ at which we will have complete information loss, i.e., all observed values map to a single bin. For each continuous variable in the state-action space, we consider a sequence of quantizers with decreasing step size $Q = \{Q_{\sigma_j}\}_ {j=0,\dots, K}$, $\sigma_j > \sigma_{j+1}$, which naturally gives a quantization sequence $\Bar{Q}$ for the entire state-action space, provided $K$ is fixed across the continuous dimensions. To simplify notation, we collapse the sub index and write $Q_{j}$ to stand for $Q_{\sigma_j}$. For more general quantization schemes, the main requirement is the decreasingly smaller reconstruction error for $Q_{j+1}$ in comparison to $Q_j$.

For an episode $E$, we compute its quantized representation in an obvious component-wise manner: 
\begin{equation} \label{eq:quantization}
E_j = \Bar{Q}_j(E)=\{(\Bar{Q}_j(s_t), \Bar{Q}_j(a_t))\}_{t \in {1,\dots,T}}    
\end{equation}
which defines a multi-resolution representation of the episode as a corresponding ordered set $\{E_j\}_{j \in \{0, \dots, K\}}$ of quantized episodes, where $\Bar{Q}$ is the vector version of quantization $Q$.

In the quantized Markov model $M_{n,j}= \Bar{Q}_j(M_n)$, which we construct from the episode $E_j$, we compute extended states using the corresponding quantized values. Hence, the extended state is $\Bar{Q}_j(S_{t,n})=(\Bar{Q}(s_t), \Bar{Q}(\alpha_{t,n}))$. Further, we define the model $\Bar{Q}_j(M_n)$ to contain probabilities $P\{a_t|\Bar{Q}_j(S_{t,n})\}$ for the \emph{original} action values. In other words, we do not rely on the reconstruction mapping $\Bar{Q}_j^{-1}$ to recover action but store the original actions explicitly. In practice, continuous action values tend to be unique and the model samples from the set of values observed after the occurrences of the corresponding extended state. Our experiments show that replaying the original actions instead of their quantized representation provides better continuity and natural true-to-the-demonstration look of the cloned behavior.

\textbf{Markov Ensemble:} Combining together stacking and multi-resolution quantization of Markov models, we obtain Markov Ensemble $\mathcal{E}$ as an array of Markov models parameterized by the model order $n$ and the quantization schema $Q_{j}$:

\begin{equation} \label{eq:ensemble}
\mathcal{E}_{N,K} = {\mathcal{M}_{i,j}}, i=0, \dots, N, j=0,\dots,K
\end{equation}

The policy defined by the ensemble (\ref{eq:ensemble}) computes each next action following Algorithm \ref{alg:markov_ensemble}. The Markov Ensemble technique, together with the policy defined by it, are our primary tools for cloning behavior from demonstrations.

Note, that with the coarsest quantization $\sigma_0$ present in the multi-resolution schema, the policy should always return an action sampled using one of the quantized models, which at the level $0$ always finds a match. Hence, such models always ``generalize'' by resorting to simple sampling of actions when no better match found in the observations. Excluding too coarse quantizers and Markov order 0 will result in executing default policy $\pi_*$ in the Algorithms  \ref{alg:markov_stack_discrete} and \ref{alg:markov_ensemble}. A possible default policy returns a random action sampled from the action space as in the provided OpenAI Gym examples. The default policy in a video game can be as simple (return an idle action) or more sophisticated and use some heuristics preventing the agent from erratic behavior.

\begin{algorithm}[tb]
   \caption{Markov Stack $\mathcal{M}$ Discrete Partial Policy}
   \label{alg:markov_stack_discrete}
\begin{algorithmic}
   \STATE {\bfseries Input:} sequence $E=\{(s_t, a_t)\}_{t \in {1,\dots,T}}$, time $t$.
   \STATE {\bfseries Output:} next action $a$
   \STATE // Note the direction of the iterations:
   \FOR{$k=n$ {\bfseries to} $0$ step $-1$} 
\STATE Form extended state $S=(s_{t},\alpha_{t,k})$.
   \IF{$S \in$ dom $M_n$}
   \STATE RETURN $M_n(S)$
   \ENDIF
   \ENDFOR
   \STATE RETURN fallback action $a = \pi_*(s_t)$
\end{algorithmic}
\end{algorithm}

\begin{algorithm}[tb]
   \caption{Markov Ensemble $\mathcal{E}_{N,K}$ Policy}
   \label{alg:markov_ensemble}
\begin{algorithmic}
   \STATE {\bfseries Input:} sequence $E=\{(s_t, a_t)\}_{t \in {1,\dots,T}}$, time $t$.
   \STATE {\bfseries Output:} next action $a$
   \STATE // Note the direction of the iterations:
   \FOR{$j=K$ {\bfseries to} $0$ step $-1$} 
   \FOR{$i=n$ {\bfseries to} $0$ step $-1$} 
\STATE Form quantized extended state 
\STATE $S=(\Bar{Q_j}(s_t), \Bar{Q_j}(\alpha_{t,i}))$.
   \IF{$S \in$ dom $M_{i, j}$}
   \STATE RETURN $M_{i, j}(S)$
   \ENDIF
   \ENDFOR
   \ENDFOR
   \STATE RETURN fallback action $a = \pi_*(s)$
\end{algorithmic}
\end{algorithm}

\textbf{Interactive Training of Markov Ensemble:} If the environment allows a human to override the currently executing policy and record new actions (demonstrations), then we can generate a sequence of demonstrations produced interactively. For each demonstration, we construct a new Markov Ensemble and add it to the sequence (stack) of already existing models. The policy based on these models consults with the latest one first. If the consulted model fails to produce an action, the next model is asked, etc. until there are no other models or one of them returns action sampled according to Algorithm \ref{alg:markov_ensemble}. Thanks to the sequential organization, the latest demonstrations take precedence of the earlier ones, allowing correcting previous mistakes or adding new behavior for the previously unobserved situations. We illustrate the logic of such an interaction with the sample git repository \cite{ibor_github_jun2019}. The computational costs for each ensemble, as already noted, is small constant $\mathcal{O}(1)$ while the overall complexity grows linearly with the number of demonstrations, allowing sufficiently long interaction of a user with the environment and training a more powerful policy. In our case studies, we show that often even a small number of strategically provided demonstrations results in a well-behaving policy.

\begin{figure}
\centering
\includegraphics[width=\columnwidth]{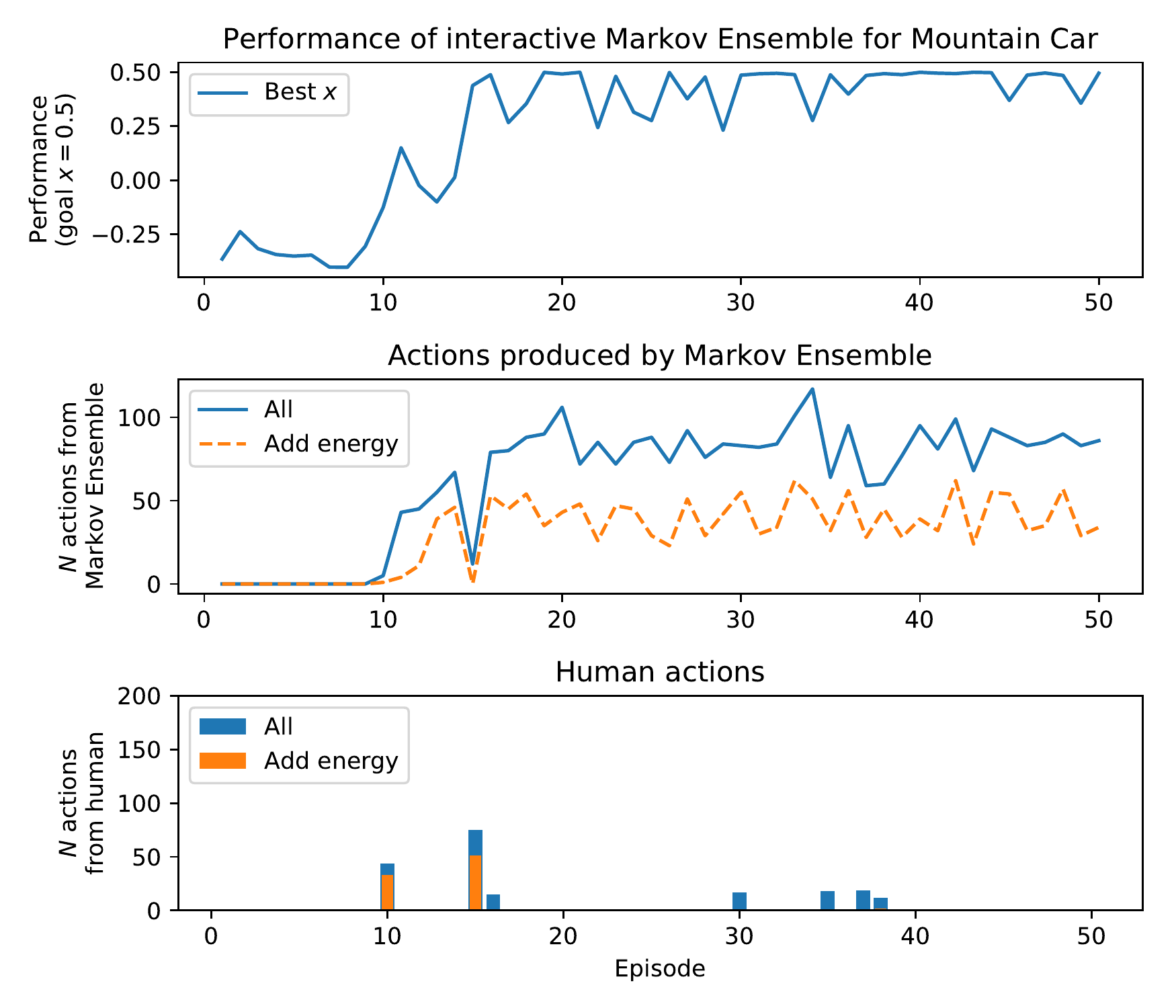}
\caption{During a number of Mountain Car episodes, a human user can take control over and provide demonstrations (bottom bar chart). The random agent is controlling the car in the first 10 episodes to establish baseline. For this environment, good actions (orange color) add energy to the car. Overall, Ensemble model becomes more ``competent'' with more demonstrations, i.e., is capable to provide next action for more observations, eventually solving most episodes.}
\label{fig:Mountain_Car}
\end{figure}



\section{Case Studies}
We consider two simple OpenIA Gym examples first and then discuss a more practical application of the proposed interactive training to a proprietary modern open-world first person shooter video game.

\textbf{OpenAI Gym, classic control}: We discuss Mountain Car and Lunar Lander to illustrate the approach with working Python code, available from the repository \cite{ibor_github_jun2019}. 
The first example, Mountain Car, is nearly trivial for a human to solve. Also, it allows the evaluation of the quality of individual actions: the ``good'' actions add mechanical energy to the system. The plot on Figure \ref{fig:Mountain_Car} shows that after just a few episodes that included some demonstrations from a human, the model can solve the environment on its own quite often. Its performance keeps improving with additional demonstrations even if some of the demonstrated actions may be sub-optimal.

Lunar Lander poses more challenges to a human. It requires some level of skills from a human player to outperform the random agent substantially. However, despite the sub-optimality of human demonstrations, they significantly increase performance over that one of the random agent.

\begin{figure}
\centering
\includegraphics[width=\columnwidth]{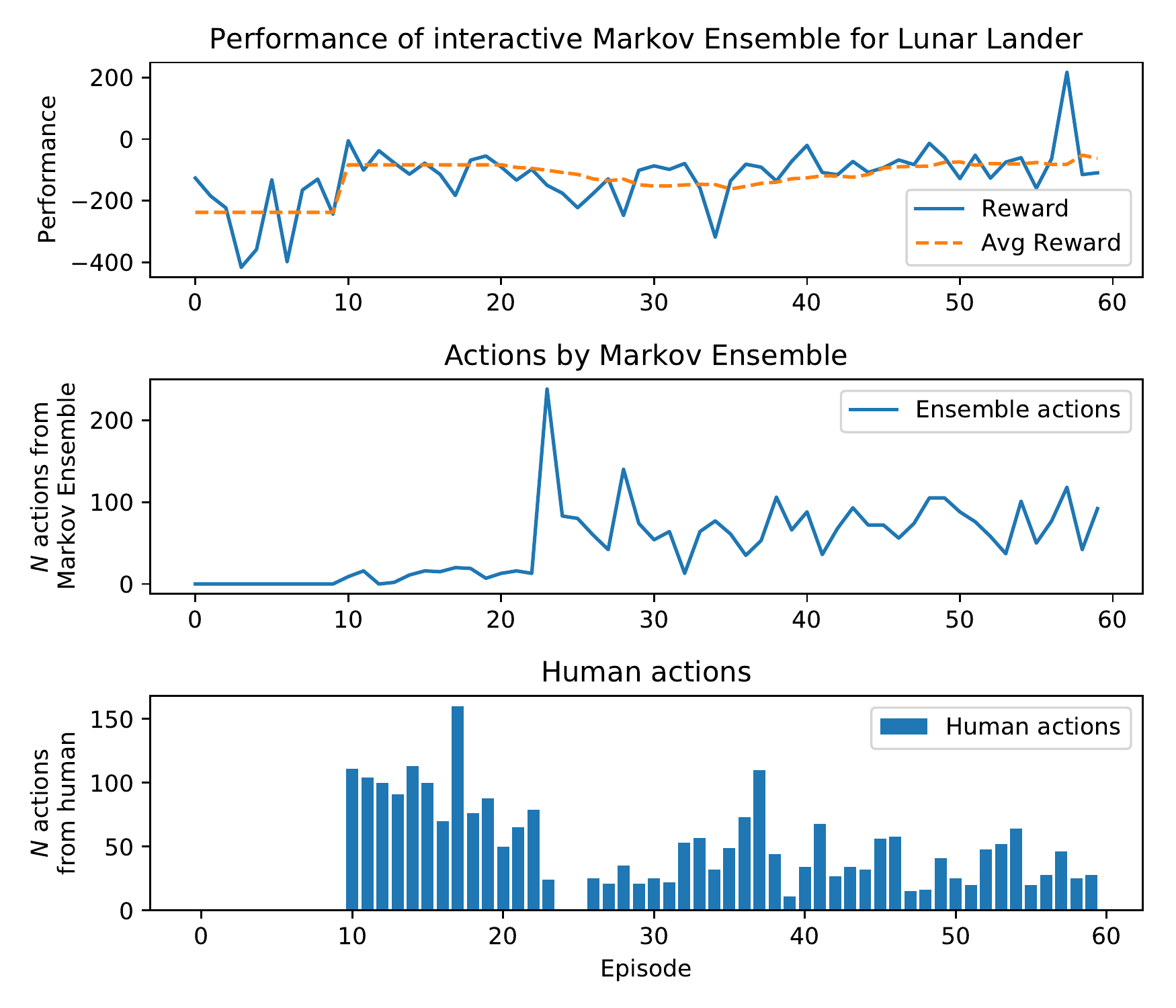}
\caption{During a number of episodes, a human can take control over Lunar Lander and provide demonstrations.  The random agent is controlling the lander in the first 10 episodes to establish baseline. Since average-skilled human actions remain sub-optimal, the resulting performance of the model doesn't reach same level of performance as in the case of Mountain Car. However, with human in the loop, there is at least one completely solved episode and the average performance of the model substantially surpasses that one of the random agent. The fact that the state-action spaces for Lunar Lander are of a higher dimensionality than that of Mountain Car results in slower learning from demonstrations, even with a human actively providing teacher inputs. }
\label{fig:Lunar_Lander}
\end{figure}

\textbf{An Open World Video Game:} For the practical application, we use an open world first-person shooter game, with OpenAI-style instrumentation for extracting game states and providing back actions. The game we explored is still in development, making learning from frame-buffer infeasible due to frequent changes of the visuals. The instrumentation, on the other hand, allows extracting relevant game state with relative ease. We convert the state into the features more suitable for training a general model. In particular, instead of the absolute coordinates and orientation of the player and non-player characters we use their relative to each other location and orientation. Additionally, we included the presence of other relevant features like line of site, ammo, health, ground speed, animation type (e.g., crouched, sprinting, jumping), collision state, etc. The quantization schemes come naturally from the gameplay design,  e.g., for the distance to the adversary we defined ranges like ``too far to aim'', ``can shoot but the damage is negligible'', etc., all the way down to ``melee weapon range'' using game tuning data and personal experience of playing the game. 

\begin{figure}
\centering
\includegraphics[width=\columnwidth]{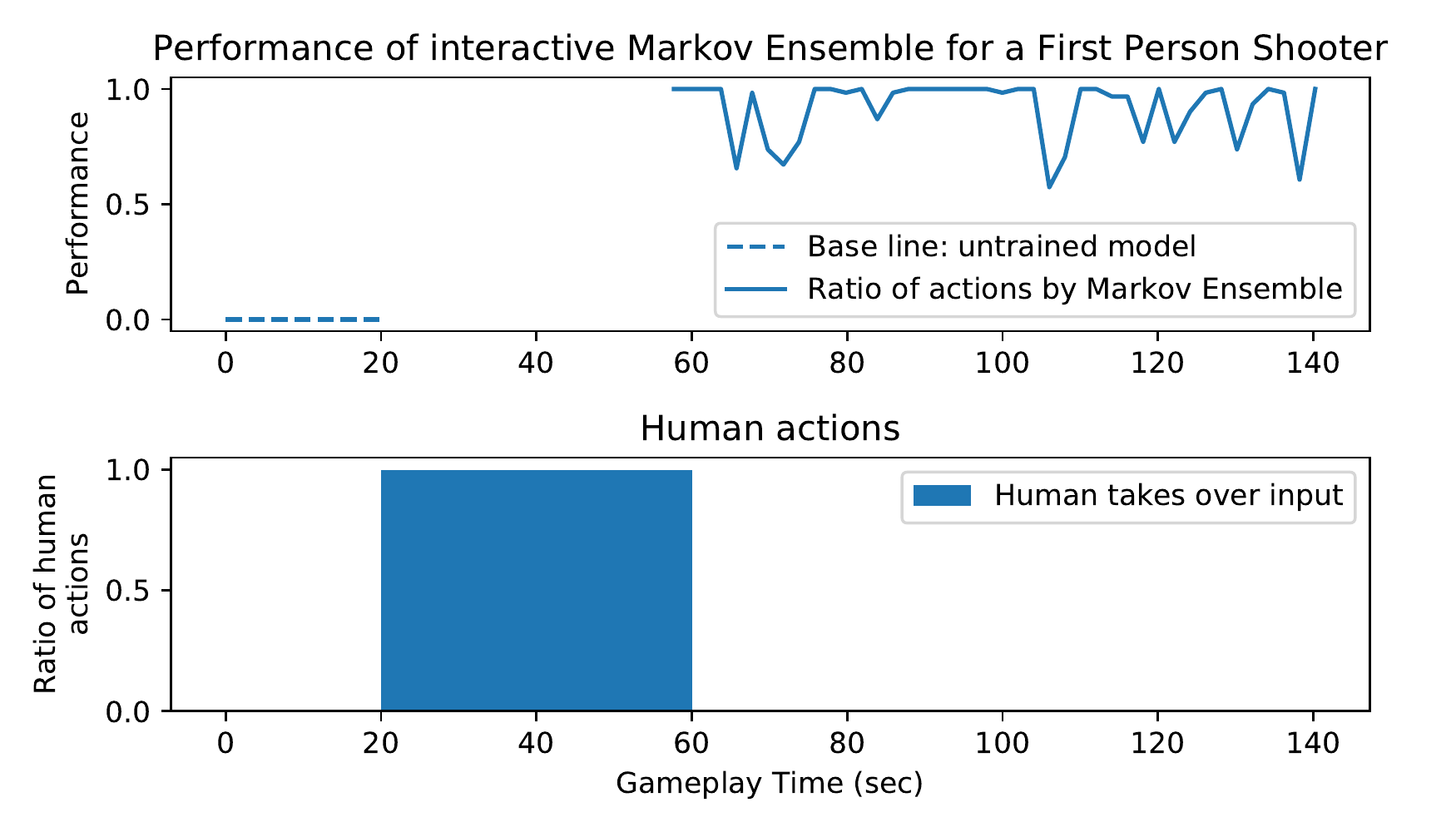}
\caption{During a session of gameplay in a First Person Shooter game where an AI agent is in control, a human user takes over to teach the model new skills, resulting in improved performance by the model. Here, the performance is the proportion of time that the underlying Markov Ensemble was able to recognize the game state and supply an action. Concretely, the AI agent learned to follow and attack the adversarial NPC using ranged weapons aggressively. Training happens between 20 and 60 seconds into the episode.}
\label{fig:Interactive_Learning_Performance}
\vskip -0.2in
\end{figure}

For the experiments reported in Figure \ref{fig:Interactive_Learning_Performance}, we trained only one type of gameplay - aggressive approach and attack of the adversary. The untrained model would not attempt to approach and attack without additional fallback logic in the control loop. After only about 40 seconds of human training, the Markov Ensemble model learns attack skills to a high degree of efficiency and eliminates the need for the additional logic most of the time. Here, efficiency refers to how frequently the Markov Ensemble is able to infer an action for the current extended state and not how competitive the performance of those actions is. Similarly, we can train an agent to negotiate contextual obstacles, use ``goodies'' (e.g., medical and ammo packs found in the environment) and interact with other types of objects in the game. 

Since loading and inference time of the stacked model grows as $\mathcal{O}(N)$ with the number of demonstrations $N$, it may be beneficial to apply DAGGER \cite{Ross2011ARO} to build a more compact aggregate model. Alternatively, we can use bootstrap to generate additional training data with no human in the loop to train an aggregate DNN model from the resulting data \cite{FromDemo}. When using ANNs, aggregated models would load and execute inference at the speeds compatible with the real-time performance of the game.

\section{Discussion and Future Work}

In this paper, we show that including a human in the learning loop can result in practical training of even such a simple model like our Markov Ensemble model. Quantization provides a simple yet efficient way of generalization for the models trained only on a limited number of human inputs. The process of having a human actively monitor the model and take over control can train practically useful models like the one we discuss for a first-person shooter game. Our future work includes meta-parameters tuning for the ensemble parameters (quantization schemes, extended state history size) for more effective training and more accurate style reproduction. Also, it will be beneficial to place the proposed techniques into a more general framework of RL and IL.

\nocite{FromDemo}
\bibliography{icml_hill}
\bibliographystyle{icml2019}





\end{document}